\documentclass{sig-alternate-2013}

\usepackage{balance}     
\usepackage{booktabs}
\usepackage{bm}
\usepackage{epstopdf}
\usepackage{graphicx}
\usepackage{flushend}	
\usepackage{url}
\usepackage{qtree} 

\makeatletter
\newif\if@restonecol
\makeatother

\usepackage[linesnumbered,ruled,vlined]{algorithm2e}

\newcommand{\bigO}[1]{\ensuremath{\mathop{}\mathopen{}\mathcal{O}\mathopen{}\left(#1\right)}}
\permission{Copyright is held by the International World Wide Web Conference Committee (IW3C2). IW3C2 reserves the right to provide a hyperlink to the author's site if the Material is used in electronic media.}
\conferenceinfo{WWW 2015,}{May 18--22, 2015, Florence, Italy.}
\copyrightetc{ACM \the\acmcopyr}
\crdata{978-1-4503-3469-3/15/05. \\
http://dx.doi.org/10.1145/2736277.2741133}

\begin{document}
\title{Describing and Understanding Neighborhood Characteristics through Online Social Media}
\numberofauthors{4} 
\author{%
\alignauthor
Mohamed Kafsi\\
\affaddr{EPFL}\\
\affaddr{Lausanne, Switzerland}\\
\email{mohamed.kafsi@epfl.ch}
\alignauthor
Henriette Cramer\\
\affaddr{Yahoo Labs}\\
\affaddr{Sunnyvale, CA, USA}\\
\email{\mbox{henriette@yahoo-inc.com}}
\and
\alignauthor Bart Thomee\\
\affaddr{Yahoo Labs}\\
\affaddr{San Francisco, CA, USA}\\
\email{\mbox{bthomee@yahoo-inc.com}}
\alignauthor David A.\ Shamma\\
\affaddr{Yahoo Labs}\\
\affaddr{San Francisco, CA, USA}\\
\email{aymans@acm.org}
}
\maketitle

\begin{abstract}
Geotagged data can be used to describe regions in the world and discover local themes. However, not all data produced within a region is necessarily specifically descriptive of that area. To surface the content that is characteristic for a region, we present the \textit{geographical hierarchy model} (GHM), a probabilistic model based on the assumption that data observed in a region is a random mixture of content that pertains to different levels of a hierarchy. We apply the GHM to a dataset of 8 million Flickr photos in order to discriminate between content (i.e.,\ tags) that specifically characterizes a region (e.g.,\ neighborhood) and content that characterizes surrounding areas or more general themes. Knowledge of the discriminative and non-discriminative terms used throughout the hierarchy enables us to quantify the uniqueness of a given region and to compare similar but distant regions. Our evaluation demonstrates that our model improves upon traditional Naive Bayes classification by 47\% and hierarchical TF-IDF by 27\%. We further highlight the differences and commonalities with human reasoning about what is locally characteristic for a neighborhood, distilled from ten interviews and a survey that covered themes such as time, events, and prior regional knowledge.
\end{abstract}

\newpage
\section{Introduction}
\noindent Finding characteristics that are specific to a geographic region is challenging because it requires local knowledge to identify what is particularly salient in that area.
Knowledge of regional characteristics becomes critical when communicating about or describing regions, for instance in the context of a mobile travel application that provides country, city and neighborhood summaries. This knowledge is especially useful when comparing regions to make geo-based recommendations: tourists visiting Singapore, for example, might be interested in exploring the Tiong Bharu neighborhood if they are aware it is known for its coffee in the same way the San Francisco Mission district is.

Photo-sharing websites, such as Instagram and Flickr, contain photos that connect geographical locations with user-generated annotations. We postulate that local knowledge can be gleaned from these geotagged photos, enabling us to discriminate between annotations (i.e.,\ tags) that specifically characterize a region (e.g.,\ neighborhood) and those that characterize surrounding areas or more general themes. This is, however, challenging because much of the data produced within a region is not necessarily specifically descriptive for that area. Is the word ``desert'' specifically descriptive of Las Vegas, or rather of the surrounding area? Can we quantify to what extent the word ``skyscraper'' is descriptive of Midtown, Manhattan, New York City or the United States as a whole?

In this paper, we propose the geographical hierarchy model (GHM), a probabilistic hierarchical model that enables us to find terms that are specifically descriptive of a region within a given hierarchy. The model is based on the assumption that the data observed in a region is a random mixture of terms generated by different levels of the hierarchy. Our model further gives insight into the diversity of local content, for example by allowing for identification of the most unique or most generic regions amongst all regions in the hierarchy. The GHM is flexible and generalizes to both balanced and unbalanced hierarchies that vary in size and number of levels. Moreover, it is able to scale up to very large datasets because its complexity scales linearly with respect to the number of regions and the number of unique terms. To evaluate our model, we compare its performance with two methods: Naive Bayes~\cite{Lewis:1998} and a hierarchical variant of TF-IDF~\cite{rattenbury2009methods}.

To investigate how well the descriptive terms surfaced by our model for a given region correspond with human descriptions, we focus on annotations of photos taken in neighborhoods of San Francisco and New York City. We apply our method to a dataset of 8 million geotagged photos described by approximately 20 million tags. We are able to associate to each neighborhood the tags that describe it specifically, and coefficients that quantify its uniqueness. This enables us to find the most unique neighborhoods in a city and to find mappings between similar neighborhoods in both cities. 

We contrast the neighborhood characteristics uncovered by our model with their human descriptions by conducting a survey and a user study. This allows us not only to assess the quality of the results found by the GHM, but mainly to understand the human reasoning about what makes a feature distinctive (or not) for a region. Beyond highlighting individual differences in people's local experiences and perceptions, we touch topics such as the importance of supporting feature understanding, and consideration of adjacency and topography through porous boundaries.

The remainder of this paper is organized as follows. We first discuss related work in Section~\ref{section:relatedwork}. We then present our geographical hierarchical model in Section~\ref{section:model}, evaluate its performance in Section~\ref{section:evaluation}, and describe the user study in Section~\ref{section:userstudy}. We finally conclude the paper and present future outlooks in Section~\ref{section:conclusions}.
\section{Related Work}
\label{section:relatedwork}
\noindent Human activities shape the perception of coherent neighborhoods, cities and regions; and vice versa. As Hillier and Vaughan~\cite{Hillier2007} pointed out, urban spatial patterns shape social patterns and activity, but in turn can also reflect them. Classic studies such as those by Milgram and Lynch~\cite{Lynch1960} on mental maps of cities and neighborhoods reflect that people's perceptions go well beyond spatial qualities. They illustrate individual differences between people, but also the effects of social processes affecting individual descriptions. People's conceptualization of region boundaries are fuzzy and can differ between individuals, even while they are still willing to make a judgement call on what does or does not belong to a geographic region~\cite{montello2003s}. Urban design is argued to affect the imageability of locales, with some points of interests for example may be well known, whereas other areas in a city provide less imageability and are not even recalled by local residents~\cite{Lynch1960}. The perceived identity of a neighborhood differs between individuals, but can relate to their social identity and can influence behavior~\cite{Stedman2002}. Various qualitative and quantitative methods, from surveys of the public to trained observation, have been employed over the years to gain such insights~\cite{montello2003s,Stedman2002,egenhofer1995naive}, with the fairly recent addition of much larger datasets of volunteered geographic information~\cite{elwood2013} and other community-generated content to benefit our understanding of human behavior and environment in specific regions. Geotagged photos and their tags are also an invaluable data source for research about social photography, recommendations, and discovery~\cite{1149949,1437389}.

We differentiate our approach from those aiming to discover new regions~\cite{Thomee:region} or redefine neighborhood boundaries~\cite{cranshaw2012livehoods} using geotagged data. Backstrom et al.~\cite{Backstrom:2008:SVS:1367497.1367546} presented a model to estimate the geographical center of a search engine query and its spatial dispersion, where the dispersion measure was used to distinguish between local and global queries. Ahern et al.~\cite{ahern2007world} used \textit{k}-means clustering to separate geotagged photos into geographic regions, where each region was described by the most representative tags. This work was followed up by Rattenbury and Naaman~\cite{rattenbury2009methods}, in which the authors proposed new methods for extracting place semantics for tags. While these prior works principally attempted to identify new regions or arbitrary spaces, we instead aim to find the unique characteristics of regions in a known hierarchy. For each region at each level in the hierarchy we aim to find a specifically descriptive set of tags, drawn from the unstructured vocabulary of community-generated annotations.

Hollenstein and Purves~\cite{hollenstein2010exploring} examined Flickr images to explore the terms used to describe city core areas. For example, they found that terms such as \texttt{downtown} are prominent in North America, \texttt{cbd} (central business district) is popular in Australia and \texttt{citycenter} is typical for Europe. The authors further manually categorized tags in the city of Z\"{u}rich according to a geographical hierarchy; even with local knowledge they found that doing so was tedious because of tag idiosyncrasies and the diversity of languages. In contrast, our method enables us to automatically obtain such classifications. Moreover, we are not just able to distinguish between local and global content, but can classify a tag according to a geographical hierarchy with an arbitrary number of levels.

Hierarchical mixture models can be used for building complex probability distributions. The underlying hierarchy, that encodes the specificity or generality of the data, might be known a priori~\cite{McCallum:1998} or may be inferred from data by assuming a specific generative process, such as the Chinese restaurant process~\cite{Blei:chinese} for Latent Dirichlet Allocation~\cite{Blei:LDA,Ahmed:tweets}, where the regions and their hierarchy are learned from data. We emphasize that, in this paper, we do not try to learn latent geographical regions, but rather aim to describe regions that are known a priori. Moreover, these regions are structured according to a known hierarchy.

\section{Geographical Hierarchy Model}
\label{section:model}
\noindent Finding terms that are specifically descriptive of a region is a challenging task. For example, the tag \texttt{paname} (nickname for Paris) is frequent across many neighborhoods of Paris but is specific to the city, rather than to any of the neighborhoods in which the photos labeled with this tag were taken. The tag \texttt{blackandwhite} is also a frequent tag, which describes a photography style rather than any particular region in the world where the tag was used. The main challenge we face is to discriminate between terms that (i) specifically describe a region, (ii) those that specifically describe a sibling, parent or child region, and (iii) those that are not descriptive for any particular region. To solve this problem, we introduce a hierarchical model that is able to isolate the terms that are specifically descriptive of a given region by considering not only the terms used within the region, but also those used elsewhere. Our model goes beyond the distinction between only global and local content~\cite{Pang2011} by probabilistically assigning terms to a particular level in the hierarchy. In this paper we present our model from a geographic perspective, even though it is generic in nature: in principle any kind of hierarchy can be used where labeled instances are initially assigned to the leaf nodes.

\subsection{Definitions}
\begin{figure}
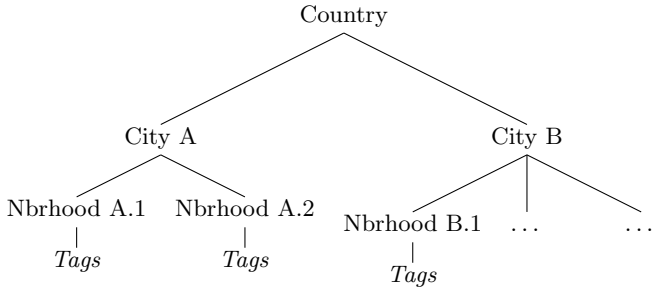

\qsetw{0.01cm}
\Tree[.Country
[.{City A} [.{Nbrhood A.1} [.{\textit{Tags}} ]] [.{Nbrhood A.2} [.{\textit{Tags}} ] ]]
[.{City B} [.{Nbrhood B.1} [.{\textit{Tags}} ] ] 
[.{$\dots$} ]
[.{$\dots$} ]]]
\caption{We represent the geographical hierarchy Country $\rightarrow$ City $\rightarrow$ Neighborhood as a geo-tree. Each node $v$ of the geo-tree is associated to a multinomial distribution over tags $\theta_v$.}
\label{figure:geotree} 
\end{figure}

\begin{description}
\item[tag] is the basic semantic unit and represents a term from a vocabulary indexed by $t \in \lbrace 1, \ldots, T \rbrace$.
\item[neighborhood] is the basic spatial unit. The semantic representation of a neighborhood is based on the collection of tags associated with the geotagged photos taken within the neighborhood. Consequently, we associate with each neighborhood $n \in \lbrace 1, \ldots, N \rbrace$, where $N$ is the number of neighborhoods, the vector $\bm{x}_{n} \in \mathbb{N}^T$, where $x_{nt}$ is the number of times tag $t$ is observed in neighborhood $n$.
\item[geo-tree] is the tree that represents the geographical hierarchy. Each node $v$ of the geo-tree is associated with a multinomial distribution $\theta_v$ such that $\theta_v(t)$ is the probability to sample the tag $t$ from node $v$. We designate the set of nodes along the path from the leaf $n$ to the root of the geo-tree as $R_n$, whose cardinality is $\vert R_n \vert$. We use the hierarchy Country $\rightarrow$ City $\rightarrow$ Neighborhood, illustrated in Figure~\ref{figure:geotree}, as the leading example throughout this paper.
\end{description}

\subsection{Model}
\noindent The principle behind our model is that the tags observed in a given node are a mixture of tags specific to the node and of tags coming from different levels in the geographic hierarchy. We represent the tags as multinomial distributions associated with nodes that are along the path from leaf to root. This model enables us to determine the tags that are specifically descriptive of a given node, as well as quantify their level of specificity or generality. The higher a node is in the tree, the more generic its associated tags are, whereas the lower a node is, the more specific its tags are. The tags associated with a node are shared by all its descendants.

We can formulate the random mixture of multinomial distributions with respect to a latent (hidden) variable 
$z \in \lbrace 1, \ldots, \vert R_n \vert \rbrace$ that indicates for each tag the level in the geo-tree from which it was sampled. For a tag $t$ observed in neighborhood $n$, $z = 1$ means that this tag $t$ was sampled from the root node corresponding to the most general distribution $\theta_{\text{root}}$, whereas $z = \vert R_n \vert$ implies that the tag $t$ was sampled from the most specific neighborhood distribution $\theta_{n}$. This is equivalent to the following generative process for the tags of neighborhood $n$: (i) randomly select a node $v$ from the path $R_n$ with probability $p(v \vert n)$, and (ii) randomly select a tag $t$ with probability $p(t \vert v)$. We suppose that tags in different neighborhoods are independent of each other, given the neighborhood in which they were observed. Consequently, we can write the probability of the tags $\bm{x}_1, \ldots, \bm{x}_n$ observed in the $N$ different neighborhoods as
\begin{equation}
\label{equation:inter}
p(\bm{x}_1, \dots, \bm{x}_n) = p(\bm{x}_1) \dots p(\bm{x}_N) .
\end{equation}
We further assume tags are independent of each other given their neighborhood, such that we can write the probability of the vector of tags $\bm{x}_n$ as:
\begin{equation}
\label{equation:intra}
p(\bm{x}_n) = \prod_{t=1}^{T} p(t \vert n)^{x_{nt}} .
\end{equation}
The probability of observing tag $t$ in neighborhood $n$ is then:
\begin{align}
\label{equation:mixture}
p(t \vert n) & = \sum_{v \in R_n} p(t \vert v) \, p(v \vert n) = \sum_{v \in R_n} \theta_{v}(t) \, p(v \vert n) ,
\end{align}
which expresses the fact that the distribution of tags in neighborhood $n$ is a random mixture over the multinomial distributions $\theta_{v}$ associated with the nodes along the path from the leaf $n$ up to the root of the geo-tree. The random mixture coefficients are the probabilities $p(v \vert n)$. By combining \eqref{equation:inter}, \eqref{equation:intra} and \eqref{equation:mixture} we obtain the log-likelihood of the data:
\begin{align}
\label{equation:likelihood}
\log p(\bm{x}_1, \dots, \bm{x}_N) &= \log \prod_{n=1}^{N} \prod_{t=1}^{T} p(t \vert n)^{x_{nt}} \nonumber \\
&= \sum_{n=1}^{N} \sum_{t =1}^{T} x_{nt} \log p(t \vert n) \nonumber \\
& = \sum_{n=1}^{N} \sum_{t =1}^{T} x_{nt} \log \sum_{v \in R_n} \theta_{v}(t) \, p(v \vert n) .
\end{align}

\paragraph*{Classification}
For a tag $t$ observed in neighborhood $n$, we can compute the posterior probability that it was generated from a given level $z$ of the geo-tree: we apply Bayes' rule to compute the posterior probability of the latent variable $z$
\begin{equation}
\label{eq:posterior}
p(z \vert t, n) = \dfrac{p(t \vert n, z) p(z \vert n)}{\sum_{z = 1}^{\vert R_n \vert} p(t \vert n, z) p(z \vert n)} .
\end{equation}
Since we assume that the distribution of tags in neighborhood $n$ is a random mixture over the distributions $\theta_{v}$ associated with nodes that forms the path $R_{n}$ from leaf $n$ up to the root of the geo-tree, the probability~\eqref{eq:posterior} is equal to
\begin{equation*}
p(v' \vert t, n) = \dfrac{\theta_{v'}(t) \, p(v' \vert n)}{ \sum_{v \in R_n} \theta_{v}(t) \, p(v \vert n)} ,
\end{equation*}
where $v'$ is the node in $R_n$ that is at level $z$. Classifying a tag $t$ observed in leaf $n$ amounts to  choosing the node $v' \in R_n$ that maximizes the posterior probability $p(v' \vert t, n)$.

\subsection{Learning}
\noindent The parameters of our model are the multinomial distributions $\theta_v$ associated with each node $v$ of our geo-tree and the mixture coefficients $p(v \vert n)$. In order to learn the model parameters that maximize the likelihood of data, given by~\eqref{equation:likelihood}, we use the Expectation-Maximization (EM) algorithm. This iterative algorithm increases the likelihood of the data by updating the model parameters in two phases: the E-phase and M-phase. The structure of our model allows us to derive closed form expressions for these updates. Due to lack of space, we omit to show our computations and refer the interested reader to Bishop~\cite{Bishop:PCML}. 
If a node in the geo-tree might contain a tag that was not observed in the training set, maximum likelihood estimates of the multinomial parameters would assign a probability of zero to such a tag. In order to assign a non-zero probability to every tag, we ``smooth'' the multinomial parameters: we assume that the distributions $\theta_v$ are drawn from a Dirichlet distribution. The Dirichlet distribution is a distribution of $T$-dimensional discrete distributions parameterized by a vector $\bm{\alpha}$ of positive reals. Its support is the closed standard $\left(T-1 \right)$ simplex, and it has the advantage of being the conjugate prior of the multinomial distribution. In other words, if the prior of a multinomial distribution is the Dirichlet distribution, the inferred distribution is a random variable distributed also as a Dirichlet conditioned on the observed tags. In order to avoid favoring one component over the others, we choose the symmetric Dirichlet distribution as a prior. We also assume a symmetric Dirichlet prior for the mixture coefficients $p(v \vert n)$.

\paragraph*{Complexity} 
Learning the parameters of our model using EM algorithm has, for each iteration, a worst case running-time complexity of $\bigO{N T D}$, where $N$ is the number of leaves of the tree, $T$ the vocabulary cardinality and $D$ the tree depth. GHM has therefore an important strength, as the time-complexity of training GHM scales linearly with respect to the number of leaves of the tree and the number of \textit{unique} tags rather than the number of tag instances. 
Furthermore, the number of iterations needed for EM to converge, when trained on the Flickr dataset introduced in Section~\ref{section:dataset}, is typically around 10.

\section{Evaluation}
\label{section:evaluation}
\noindent 
In this section, we apply our model to a large collection of geotagged Flickr photos taken in neighborhoods of San Francisco and New York City. 
The quality of the descriptive tags found by the GHM strongly supports the validity of our approach, which we further confirm using the results 
of the user study in Section~\ref{section:userstudy}: we approximate the probability that GHM classifies a tag ``correctly" by the average 
number of times its classification matches the experts' classification. Moreover, the GHM allows us to quantify the uniqueness of these neighborhoods and to obtain a mapping between neighborhoods in different cities (Section~\ref{section:classification}), enabling us to answer questions such as ``How unique is the Presidio neighborhood in San Francisco?'' or ``How similar is the Mission in San Francisco to East Village in New York City?''. Finally, we compare the performance of our model with the performance of other methods in classifying data generated according to a given hierarchy (Section~\ref{section:synthetic}).

\subsection{Dataset and classification}
\label{section:classification}
\noindent We now apply our model to a large dataset of user-generated content to surface those terms considered to be descriptive for different regions as seen through the eyes of the public.

\subsubsection*{Flickr dataset}
\label{section:dataset}
\noindent Describing geographical areas necessitates a dataset that associates rich descriptors with locations. Flickr provides an ample collection of geotagged photos and their associated user-generated tags. We couple geotagged photos with neighborhood data from the city planning departments of San Francisco\footnote{\url{https://data.sfgov.org} (Accessed 11/2014)} and New York City\footnote{\url{http://nyc.gov} (Accessed 11/2014)}, and focus on the neighborhoods of San Francisco (37 neighborhoods) and Manhattan (28 neighborhoods). Flickr associates an accuracy level with each geotagged photo that ranges from 1 (world level) to 16 (street level). In order to correctly map photos to neighborhoods, we only focus on photos whose accuracy level exceeds neighborhood level. We acquired a large sample of geotagged photos taken in San Francisco (4.4 million photos) and Manhattan (3.7 million photos) from the period 2006 to 2013. We preprocessed the tags associated with these photos by first filtering them using a basic stoplist of numbers, camera brands (e.g.\ Nikon, Canon), and application names (e.g\ Flickr, Instagram), and then by stemming them using the Lancaster\footnote{\url{http://comp.lancs.ac.uk/computing/research/stemming/}(Accessed 11/2014)} method. We further removed esoteric tags that were only used by a very small subset of people (less than 10). To limit the influence of prolific users we finally ensured no user can contribute the same tag to a neighborhood more than once a day. The resulting dataset contains around 20 million tags, of which 7,936 unique tags, where each tag instance is assigned to a neighborhood. These tags form the vocabulary we use for describing and comparing different regions. The geo-tree we build from this dataset has 3 levels: level one with a single root node that corresponds to the United States, level two composed of two nodes that correspond to San Francisco and Manhattan, and level three composed of 65 leaves that correspond to the neighborhoods of these cities.  


\subsubsection*{Classifying tags}
\noindent We applied our model to the Flickr dataset in order to find tags that are specifically descriptive of a region. We show the results for two neighborhoods in San Francisco---Mission and Golden Gate Park--- and two neighborhoods in Manhattan---Battery Park and Midtown---in particular. In Table~\ref{tab:TopTags} we show the top 10 \textit{most likely} tags for each of these four neighborhoods, where each tag is further classified to the \textit{most likely} level that have generated it. Effectively, each tag observed in a neighborhood is ranked according to the probability $p(t \vert n)$. Then, we apply~\eqref{eq:posterior} to compute the posterior probability of the latent variable $p(z \lvert t,n)$ and classify the tag accordingly. Recall that the value of the latent variable $z$ represents the level of the geo-tree from which it was sampled. For example, a tag observed in neighborhood $n$ is assigned to the neighborhood level if the most likely distribution from which it was sampled is the neighborhood distribution $\theta_n$. We consider such a tag as being specifically descriptive of neighborhood $n$.

\begin{table}
\centering
\begin{small}
\begin{tabular}{l l | l l}
\toprule
\textbf{Mission} & \textbf{GG Park} & \textbf{Battery Park} &
\textbf{Midtown} \\ 
\midrule
california \ddag & california \ddag & newyork \ddag & newyork \ddag \\ 
\textbf{mission} & \textbf{flower} & manhattan \ddag & manhattan \ddag \\ 
sf \ddag & \textbf{park} & \textit{usa} \dag & \textit{usa} \dag \\ 
\textit{usa} \dag & sf \ddag & \textbf{wtc} & \textbf{midtown} \\ 
\textbf{graffiti} & \textit{usa} \dag & \textbf{brooklyn} & \textbf{skyscraper} \\ 
\textbf{art} & \textbf{museum} & \textbf{downtown} & \textbf{timessquare} \\ 
\textbf{mural} & \textbf{tree} & \textbf{bridge} & \textit{light} \dag \\
\textbf{valencia} & \textbf{ggpark} & gotham \ddag & \textbf{moma} \\
\textbf{food} & \textbf{deyoung} & \textbf{memorial} & \textbf{broadway} \\ 
car \ddag & architecture \ddag & \textit{light} \dag & \textbf{rockefeller} \\
\bottomrule
\end{tabular}
\end{small}
\caption{For each neighborhood, we show the $10$ most likely tags ranked according to their probability $p(t \vert n)$. We use the posterior probability $p(z \vert t,n)$ in order to assign each tag to the distribution that maximizes this posterior probability: country (\dag), city (\ddag) or neighborhood (\textbf{bold}).}
\label{tab:TopTags}
\end{table} 

Despite the fact that the tag \texttt{california} is the most likely (frequent) tag in both Mission and Golden Gate Park, it is not assigned to the neighborhood distribution but rather to the city distribution (we presume, if we were to add a new State level to the geo-tree, that the tag \texttt{california} would most likely be assigned to it). This confirms the notion that the most frequent tag in a neighborhood does not necessarily describe it specifically. We see that \texttt{architecture} is applicable not only to the buildings in Golden Gate Park---notably the De Young---but it also is a descriptor for San Francisco in general. Our method is able to discriminate between frequent and specifically descriptive tags, whereas a naive approach would still consider a very frequent tag in a neighborhood as being descriptive. The same observation is valid for the tags \texttt{usa} and \texttt{light}, which are tags that are too general and therefore not specifically descriptive of a neighborhood. The tag \texttt{car} may seem misclassified at the city level in San Francisco, until one realizes the city is well known for its iconic cable cars. Considering that the tags \texttt{art}, \texttt{mural} and \texttt{graffiti} are classified as being specific to Mission is not surprising, because this neighborhood is famous for its street art scene. The most probable tags that are specifically descriptive of Midtown in Manhattan include popular commercial zones, such as Rockefeller Center, Times Square and Broadway, as well as the museum of modern art (MoMa). The tag \texttt{gotham}, one of the most observed tags in Battery Park, is assigned to city level, which is not unexpected given that Gotham is one of New York City's nicknames.

\subsubsection*{Neighborhood uniqueness}
\begin{figure}
\begin{center}
\includegraphics[width=\columnwidth]{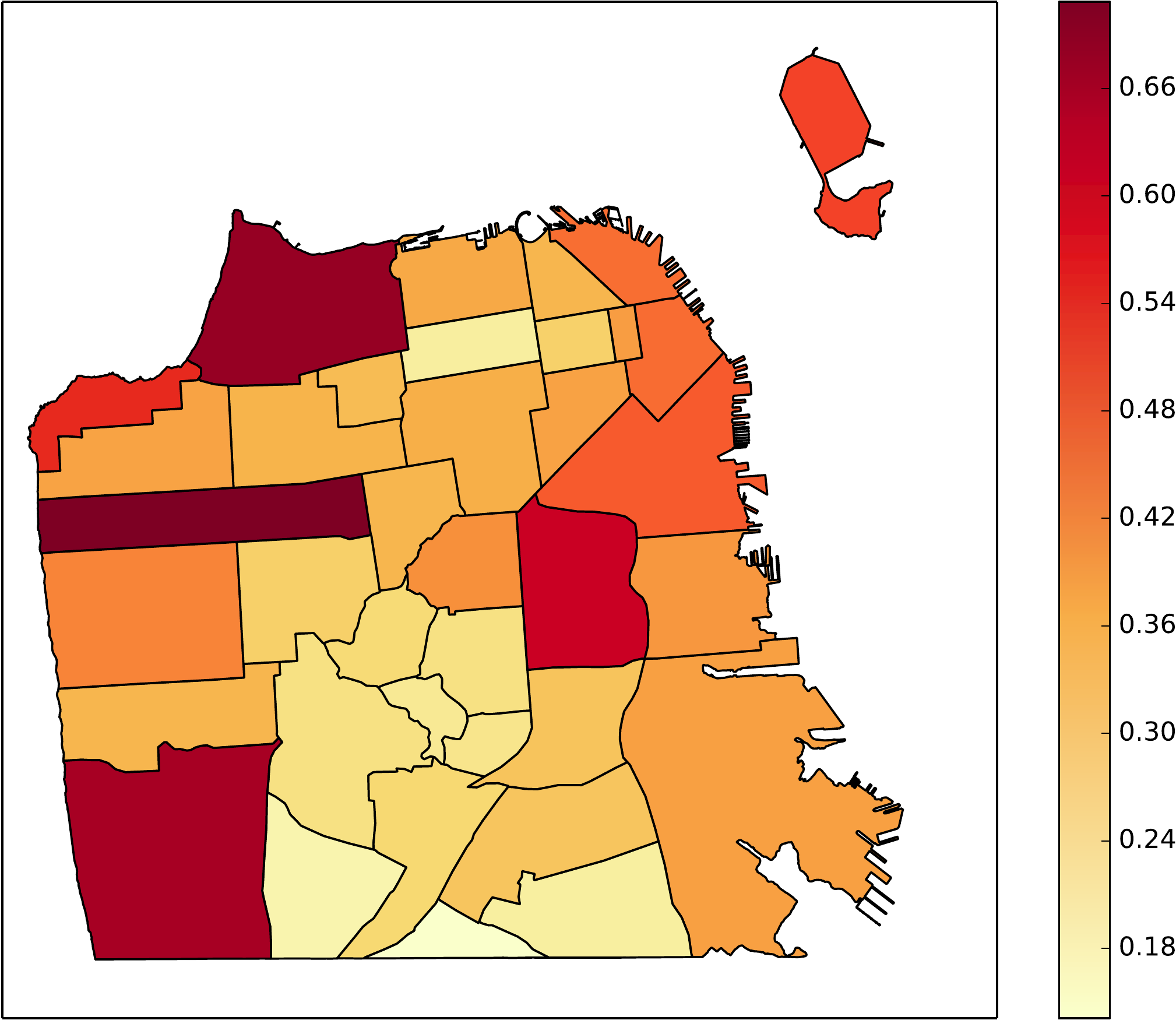}
\caption{Neighborhoods of San Francisco colored according to their local mixture coefficient $p(z \vert n)$, where $z = \vert R_n \vert$. A darker color indicates a larger local mixture coefficient (`uniqueness').}
\label{figure:sf}
\end{center}
\vspace{-6pt}
\end{figure}

\noindent \textit{If you are interested in visiting the most unique neighborhoods in San Francisco, which ones would you choose?} With our framework, we can quantify the uniqueness of neighborhood $n$ by using the probability $p(z \vert n)$ of sampling local tags, where $z = \vert R_n \vert$. 
In fact, a high probability indicates that we sample often from the distribution of local tags $\theta_n$, and can therefore be interpreted as indicator of a more unique local character. We show a map of the city of San Francisco in Figure~\ref{figure:sf}, in which each neighborhood is colored proportionally to its local-mixture coefficient. The darker the color is, the more unique the personality of the neighborhood is. The four most unique neighborhoods in San Francisco are Golden Gate Park (0.70), Presidio (0.67), Lakeshore (0.65) and Mission (0.60), which is not surprising if you know these neighborhoods. The Golden Gate park is the largest urban park in San Francisco, famous for its museums, gardens, lakes, windmills and beaches. The Presidio is a national park and former military base known for its forests, scenic points overlooking the Golden Gate Bridge and the San Francisco Bay. Lakeshore is known for its beaches, the San Francisco zoo and also for San Francisco State University. The Mission is famous for its food, arts, graffiti, and festivals.

\subsubsection*{Mapping neighborhoods between cities}
\noindent \textit{Given a neighborhood in San Francisco, what is the most similar neighborhood in Manhattan?} To answer such a question, we can use our framework to find a mapping between similar neighborhoods that are in different cities and even different countries. Recall that each neighborhood $n$ is described by its local distribution $\theta_n$. In order to compare two neighborhoods $n$ and $n'$, we compute the cosine similarity between their respective local distributions $\theta_n$ and $\theta_{n'}$, given by:
\begin{equation}
\label{eq:cosim}
\text{sim}(\theta_n,\theta_{n'}) = 
\frac{\sum_{t=1}^{T}\theta_n(t) \,\theta_{n'}(t)}{\sqrt{\sum_{t=1}^{T}\theta_n(t)^2} \sqrt{\sum_{t=1}^{T}\theta_{n'}(t)^2}} .
\end{equation}
The similarity range is $[0,1]$, with $\text{sim}(\theta_n,\theta_n')=1$ if and only if $\theta_n=\theta_n'$. Table~\ref{table:mapping} shows the mapping from six San Francisco neighborhoods to the most similar neighborhoods in Manhattan respectively. We also include the second most similar neighborhood when the similarities are very close. In order to give some intuition about
the mapping obtained, we also show the top five common local tags obtained by ranking the tags $t$ according to the product of tag probabilities $\theta_n(t) \, \theta_{n'}(t)$. For example, the East Village in Manhattan is mapped to Mission in San Francisco; the strongest characteristics they share are graffiti/murals, food, restaurants and bars. Moreover, despite the major differences between San Francisco and Manhattan, their Chinatowns are mapped to each other and exhibit a highly similar distribution of local tags (cosine similarity of 0.85). Finally, it is not surprising that Treasure Island for San Francisco and Roosevelt Island for Manhattan are mapped to each other, since both are small islands located close to each city. We however emphasize that the top \textit{common} local tags between neighborhoods are not necessarily most descriptive for the \textit{individual} neighborhoods. The top tags may rather provide a shallow description (e.g.\ dragons in Chinatown, boats for island) and are useful to gain some insight into the mapping obtained, but we are aware that every Chinatown is unique and not interchangeable with another one. 

\begin{table*}
\centering
\begin{small}
\begin{tabular}{l | l | l}
\toprule
\textbf{San Francisco} & \textbf{Manhattan} & \textbf{Top common
local tags} \\
\midrule
Mission & East Village (0.23) & \texttt{graffiti}, \texttt{food}, \texttt{restaurant}, \texttt{mural}, \texttt{bar} \\ 
Golden Gate Park & Washington Heights (0.26), Upper West Side (0.22) & \texttt{park}, \texttt{museum}, \texttt{nature}, \texttt{flower}, \texttt{bird} \\
Financial District & Battery Park (0.29), Midtown Manhattan (0.27) & \texttt{downtown}, \texttt{building}, \texttt{skyscraper}, \texttt{city}, \texttt{street} \\ 
Treasure Island & Roosevelt Island (0.38) & \texttt{bridge}, \texttt{island}, \texttt{water}, \texttt{skylines}, \texttt{boat} \\
Chinatown & Chinatown (0.85) & \texttt{chinatown}, \texttt{chinese}, \texttt{downtown}, \texttt{dragons}, \texttt{lantern} \\
Castro & West Village (0.06) & \texttt{park}, \texttt{gay}, \texttt{halloween}, \texttt{pride}, \texttt{bar} \\
\bottomrule
\end{tabular}
\end{small}
\caption{Mapping from San Francisco neighborhoods to the most similar ones in Manhattan. For each neighborhood pair $n$ and $n'$, we give the cosine similarity between their local distributions $\theta_n$ and $\theta_{n'}$, and list the top ``common'' local tags ranked according to the product of tag probabilities $\theta_n(t) \, \theta_{n'}(t)$.}
\label{table:mapping}
\end{table*}

\subsection{Experimental evaluation}
\label{section:synthetic}
\noindent Evaluating models such as the GHM on user-generated content is hard because of the absence of ground truth. There is no dataset that associates, with objectivity, regions to specifically descriptive terms, or assigns terms to levels in a geographic hierarchy. This is due to the intrinsic subjectivity and vagueness of the human conception of regions and their descriptions~\cite{hollenstein2010exploring}. In the absence of ground truth, a classic approach is to generate a dataset with known ground truth, and then use it to evaluate the performance of different classifiers. We follow the generative process presented in Algorithm~\ref{alg:data} using the geo-tree (3 levels, 68 nodes) built from the dataset presented in Section~\ref{section:dataset}. For each node $v$ in our geo-tree, we sample the distribution of tags $\theta_v$ from the symmetric Dirichlet distribution $Dir(\alpha)$. We set $\alpha = 0.1$ to favor sparse distributions $\theta_v$. If the node is a leaf, i.e.\ it represents a neighborhood, we also sample the mixture coefficient $p(z \vert v)$ from a symmetric Dirichlet distribution $Dir(\beta)$. We set $\beta = 1.0$ to sample the mixture coefficients uniformly over the simplex and have well-balanced distributions. Now that we have the different distributions that describe the geo-tree, we can start generating the tags in each neighborhood. We vary the number of samples per neighborhood in order to reproduce a realistic dataset that might be very unbalanced: data is sparse for some neighborhoods while very dense for some others. For each neighborhood $n$, we sample uniformly the continuous random variable $\gamma$, which represents the order of the number of tags in neighborhood $n$, from the interval $\left[3,6 \right]$.  The endpoints of the support of $\gamma$ are based on the the minimum and maximum values observed in the Flickr dataset presented in Section~\ref{section:dataset}, such that an expected number of $18.8 \times 10^6$  tags will be generated, similar to the quantity of tags available in the Flickr dataset. Once the number of tags $\nu$ to be generated is fixed, we sample, for each iteration, a level of the geo-tree and then a tag from the distribution associated to the corresponding node.
\begin{algorithm}
\KwIn{Geo-tree V, neighborhoods $n \in \lbrace 1, \ldots, N \rbrace$, tags $t \in \lbrace 1, \ldots, T \rbrace$, hyper-parameters $\alpha, \beta$.}
\KwOut{Tags observed in each neighborhood $\bm{x}_1, \ldots, \bm{x}_N$.}
\For{$v \in V$}{%
Sample distribution $\theta_v \sim Dir(\alpha)$\;
\If{v is a leaf}{Sample mixture coefficients $p(z \vert v) \sim Dir(\beta)$\;}
}
\For{$n \in \lbrace 1, \ldots, N \rbrace$}{%
Initialize $\bm{x}_n = \bm{0}$\;
Sample order $\gamma \sim \mathcal{U} \left[3,6 \right]$\; 
Set $\nu = \lfloor 2 \times 10^{\gamma} \rfloor$\;
\While{$ \sum_t x_{nt} \leq  \nu $}{%
Sample tree level $z \sim p(v \vert n)$\;
Sample tag $t \sim p(t \vert z,n)$\;
Increment tag count $x_{nt} \gets x_{nt} + 1$\;	
}
}
\caption{Generating tags in neighborhoods}
\label{alg:data}
\end{algorithm}

We can now assess the performance of a model by quantifying its ability to correctly predict the level in the geo-tree from which a tag observed in a neighborhood was sampled. In addition to our model, we consider the following methods:
\begin{description}
\item[Naive Bayes (NB)] is a simple yet core technique in information retrieval~\cite{Lewis:1998}. Under this model, we assume that the tags observed in a class (node) are sampled independently from a multinomial distribution. Each class is therefore described by a multinomial distribution learnt from the count of tags that are observed in that class. However, since we do not use the class membership to train our methods, we assign a tag $t$, observed in neighborhood $n$, to all the classes (nodes) along the the path $R_n$, which amounts to having a uniform prior over the classes. 

\item[Hierarchical TF-IDF (HT)] is a variant of TF-IDF that incorporates the knowledge of the geographical hierarchy. This variant was used in the TagMaps method~\cite{rattenbury2009methods} to find tags that are specific to a region at a given geographical level. The method assigns a higher weight to tags that are frequent within a region (node) compared to the other regions at the same level in the hierarchy. We are able to represent each node with a normalized vector in which each tag $t$ has a weight that encodes its descriptiveness.
\end{description}

For the classification, we map a tag $t$ observed in the leaf $n$ to the level $\hat{z}$ that maximizes the probability $p(z \vert t, n)$ (NB and GHM), or the tag weight (HT). Using our ground truth, we can then approximate the probability of correct classification $p(\hat{z} = z)$ by the proportion of tags that were correctly classified. In our evaluation, we repeat the following process 1000 times: we first generate a dataset, hold out $10\%$ of the data for test purposes and train the model on the remaining $90\%$. For fair comparison, we initialize and smooth the parameters of each method similarly. Then, we measure the classification performance of each method. The final results, shown in Table~\ref{tab:pref}, are therefore obtained by averaging the performance of each method over 1000 different datasets. 

\begin{table}
\begin{center}
\begin{tabular}{ r | c }
\toprule
 & Classification Accuracy (std) \\ 
\midrule
Random & 0.33 (0.00) \\ 
NB & 0.51 (0.02) \\
HT & 0.59 (0.02)\\
\midrule
GHM & \textbf{0.75} (0.01) \\ 
\bottomrule
\end{tabular}
\caption{The average classification accuracy is computed, for each method, over 1000 generated datasets. We also indicate this accuracy if we classify tags uniformly at random.}
\label{tab:pref}
\end{center}
\end{table}

Our GHM model is the most accurate at classifying the tags to the correct level, greatly outperforming NB by 47\% and HT by 27\%. Even though both GHM and HT take advantage of the geographical hierarchy in order to classify the tags, the probabilistic nature of GHM enables a more resilient hierarchical clustering of the data, while the heuristic approach of HT suffers from overfitting. For example, if the number of samples available for a neighborhood is low, HT might overfit the training data by declaring a frequent tag as being characteristic, although not enough samples are available to conclude this. This is not case for GHM, because the assumptions of random mixture enable us to obtain a resilient estimate of the distributions, which declare a tag as characteristic of a given level only if it has enough evidence for it. This observation is strengthened if we choose the maximum order $\gamma$ of the number of tags per node to be $4$ instead of $6$: the classification accuracy of GHM decreases by $5\%$ only (0.71), whereas the performance of HT decreases by $13\%$ (0.51). Taken together, these results suggest that, if the data observed in a neighborhood is a mixture of data generated from different levels of a hierarchy that encodes the specificity/generality of the data, our method will be successfully able to accurately associate a tag to the level from which it was generated.

\section{Perception focused user study}
\label{section:userstudy}
\noindent The results produced by our model might not necessarily be intuitive nor expected, especially in the light of people's differing individual geographic perspectives~\cite{montello2003s,Stedman2002,egenhofer1995naive}. Trying to objectively evaluate these, without taking into account human subjectivity and prior knowledge, could be misleading. For the Castro neighborhood in San Francisco, for example, the GHM classified the tag \texttt{milk} as specifically descriptive. Someone who is not familiar with Harvey Milk, the first openly gay person to be elected to public office in California and who used to live in the Castro, would most probably not relate this tag to the neighborhood.  We thus need to understand the correspondences and gaps between our model's results and human reasoning about regions. 

We conduct a user-focused study to explore the premise that the posterior probability of a tag being sampled from the distribution associated with a region is indicative of the canonical descriptiveness of this region. We further aim to identify potential challenges in user-facing applications of the model, and to uncover potential extensions to our model. We held ten interviews and conducted a survey with local residents of the San Francisco Bay Area focusing on their reactions to the tags that our model surfaced. To assess the performance of our model while reducing the bias of subjectivity, we used the results of our user survey to obtain the human classification of tags to nodes in the hierarchy, allowing us to approximate the probability that our model classified a tag correctly by the average number of times it corresponded with human classification. We use the interviews to understand the reasons behind matches and mismatches.

\subsection{Interview and survey methodology}
\noindent Interviewees and survey respondents reacted to a collection of 32 tags per neighborhood. To ensure a certain diversity among the tags presented to the users, we select randomly a subset of tags that are classified by the GHM as being descriptive of (i) neighborhood level (e.g.,\ \texttt{graffiti}), (ii) city level (e.g.,\ \texttt{nyc}), (iii) country level (e.g.,\ \texttt{usa}), or (iv) another neighborhood (e.g.,\ \texttt{mission} for the Castro neighborhood). We selected these tags randomly, with the probability of choosing a given tag $t$ proportional to the probability $p(t \vert n)$. The survey aims to highlight to which extent locals' perspectives match the results produced by our model, while the interviews highlight the reasons why and allow us to understand better the human perception of descriptiveness.

\paragraph*{Interview procedure}
Our semi-structured interviews focused on how people describe neighborhoods, and investigated their reasoning behind the level of specificity associated to 
a tag in a given neighborhood. Each one-to-one interview lasted 25--45 minutes. To ensure a wide range of (former) local to newcomer perspectives, we interviewed a total of 10 people (5F, 5M; ages 26--62, $\mu=37$, $\sigma=12$) that (had) lived in the San Francisco Bay Area from two months to 62 years. Three of the participants worked in the technology industry, two were students, one was a real estate agent, one a building manager and one a photographer. Each interview covered three different neighborhoods chosen by the participant out of 11 well-known San Francisco neighborhoods. However, one participant described only one neighborhood (due to time constraints), while another participant described four of them. Our interviews addressed:
\begin{enumerate}
\item Participants' characterization of the neighborhoods using their own words to get an understanding of the factors that are important to them.
\item Their considerations in whether a tag is perceived as specifically descriptive or not. Interviewees were first asked to classify the 32 tags presented to them as (not) specifically descriptive for the neighborhood, and explain the reasons. Then, they were shown the subset of tags that were classified by our model as specifically descriptive.
\end{enumerate}

We emphasize the fact that the participants were not told about our model, nor that the terms presented to them were actually Flickr tags. This provided broader insight into the factors that led them to classify terms as (not) specifically descriptive of a neighborhood, and allowed for identification of factors not yet addressed by the model, without biasing their judgment towards the assumptions we made (i.e., hierarchical mixture of tags). The interviews were recorded, and transcriptions were iteratively analyzed, focusing on the identification of themes in interviewees' reasoning affecting classifications of tags as (non-)descriptive. 

\paragraph*{Survey procedure}
A total of 22 San Francisco Bay Area residents (5F, 17M; ages 22--39, $\mu=33$, $\sigma=4.8$), who had lived there for an average of 5.6 years ($\sigma=4.2$), participated in our survey about San Francisco and three of its neighborhoods (Mission, Castro and North Beach). Of these 22 respondents, 18 provided tag classifications for the Mission neighborhood, 12 for the Castro and 11 for North Beach. This resulted in 1291 tag classifications, of which 561 for the Mission, 381 for the Castro and 349 for North Beach. The survey asked respondents to describe each neighborhood with their own words using open text fields, and then to classify the 32 tags presented to them as descriptive for a given neighborhood, for a higher level (the city or country), or for another neighborhood. They could also indicate if they did not find the tag descriptive for any level, or did not understand its meaning.

\subsection{Results}
\noindent  In this section, we present the results of the survey, and provide examples from the interviews to illuminate the reasoning processes on whether a tag is descriptive for a specific region. We compare the tag classifications provided by the GHM with those supplied by the participants, and we specifically focus on disagreements between neighborhood-level tag classifications in order to identify challenges in human interpretation of modeling results, and potential extensions to our model.

\begin{table*}
\centering
\begin{small}
\begin{tabular}{l | rr | rr | rr} 
\toprule
& \multicolumn{2}{c|}{\textbf{Mission}} 
& \multicolumn{2}{c|}{\textbf{Castro}} & \multicolumn{2}{c}{\textbf{North Beach}} \\
\midrule
Neighborhood-level tag count & 17 & & 17 & & 17 & \\
Neighborhood-level tag classifications & 287 & (100\%) & 201 & (100\%) & 185 & (100\%) \\ \midrule
Tags not understood & 22 & (8\%) & 3 & (1\%) & 0 & (0\%)\\
Tags seen as non-descriptive for any level & 116 & (40\%) & 50 & (25\%) & 62 & (33\%)\\
(Part of) neighborhood & 109 & (39\%) & 67 & (33\%) & 37 & (20\%)\\
Higher-level node (CA/USA) & 24 & (8\%) & 38 & (19\%) & 7 & (4\%)\\
Other neighborhood & 16 & (5\%) & 43 & (22\%) & 79 & (43\%)\\
\bottomrule
\end{tabular}
\end{small}
\caption{The distribution of answers given by survey participants. The (rounded) percentages are computed with respect to the total number of answers given. }
\label{tab:tagclass}
\end{table*} 

\begin{table*}
\centering
\begin{small}
\begin{tabular}{l | lr | lr | lr} 
\toprule
& \multicolumn{2}{c|}{\textbf{Mission}} 
& \multicolumn{2}{c|}{\textbf{Castro}} & \multicolumn{2}{c}{\textbf{North Beach}} \\
\midrule
Tag count & 32 & & 32 & & 32 & \\
Tag classications & 561 & & 381 & & 349 & \\
\midrule
GHM alignment & 0.84 && 0.81 && 0.66 \\
Misaligned tag count & 5 && 6 && 11 \\
Misaligned tags & \texttt{night}, \texttt{coffee}, \texttt{sidewalk} && \texttt{mission}, \texttt{streetcar}, \texttt{dolorespark} && \texttt{embarcadero}, \texttt{bridge}, \texttt{water}\\
& \texttt{cat}, \texttt{brannan} && \texttt{church}, \texttt{sign}, \texttt{night} && \texttt{alcatraz}, \texttt{seal}, \texttt{sea}, \texttt{crab}, \\ &&&&& \texttt{wharf}, \texttt{boat}, \texttt{bay}, \texttt{pier} \\
\bottomrule
\end{tabular}
\end{small}
\caption{Alignment of GHM tag assignments with the majority of survey respondents' assignment for each survey neighborhood. Misalignment would for example be a tag classified as neighborhood level by the GHM while the majority of survey respondents had assigned it to another neighborhood or another level.}
\label{tab:tagalignment}
\end{table*}

\subsubsection{Participant and model congruency}
\noindent The model's premise that locally frequent content is not necessarily specific to that locale was strongly supported by the interviews and the survey. None of the participants classified all of the tags that occurred \emph{frequently} in a neighborhood as specifically \textit{descriptive} of the given neighborhood. This supports the results of the GHM in classifications of very frequent but wide-spread tags as not being specifically descriptive of the neighborhood. Without prompting, interviewees mentioned terms as being too generic or specific for a given neighborhood. For example, one interviewee (F 28), when describing the Western Addition neighborhood, picked \texttt{haight} as a descriptive tag, ``because there's Haight Street in this neighborhood'', but not the tag \texttt{streets}, as ``there's streets everywhere''. Similar interview examples included: ``\texttt{california} or \texttt{usa} is a generic, or general term'' (F 28), ``I don't think of ever describing Golden Gate Park as in the USA\@. Unless I'm somewhere far away, but then I wouldn't even say USA, I would say California or San Francisco" (F 41). This result implies that participants tend to classify tags according to a geographical hierarchy, which supports the validity of the assumptions we make about the hierarchy of tags: tag specificity/generality depends on the hierarchical level from which it was sampled.

We are aware that people will not agree with every classification made by our model. Tags classified by the GHM as specifically descriptive of a neighborhood, were not necessarily perceived as such by all respondents; variations occurred between neighborhoods and between participants. As a consequence, evaluating the results of an aggregate model `objectively', as if there were only a single correct representation of a neighborhood, is difficult. However, to place the the GHM classification into context with human classification, we use the results of our survey to reduce subjective biases: we obtain a majority vote human classification by assigning each tag to the class that users have chosen most often. We then approximate the probability that the GHM classifies a tag `correctly' by the average number of times its classification matches this human majority classification. For most tags the majority assignment is aligned with the assessment of the model (Table~\ref{tab:tagalignment}). We obtained an average classification correspondence of 0.77, with a highest classification accuracy of 0.84 for the Mission neighborhood. The alignment between the model and human classification for the Castro neighborhood was 0.81. Alignment was lowest for the North Beach neighborhood with a correspondence of 0.66, mainly caused by tag classifications as `another neighborhood' (Table~\ref{tab:tagclass}). Such mismatches occur for a multitude of reasons. First of all, as a very basic requirement, participants have to understand what a tag refers to, before they can assign it to a specific level. For example, in the survey, 8\% of the answers given for the Mission neighborhood were ``I don't know what this is'' (Table~\ref{tab:tagclass}). This issue occurred less for the selection of tags for the Castro (1\%) and North Beach (0\%). The terms that users understood were necessarily perceived as either descriptive (i.e.\ assignable to a level in the geographical hierarchy) or non-descriptive (i.e.\ not belonging to any level in the hierarchy). The proportion of individual answers that are ``non-descriptive" is around $34\%$, which included answers to tags such as \texttt{cat}, \texttt{sticker} and \texttt{wall} for the Mission. The fact that these tags indeed describe content that occurs frequently in the Mission doesn't imply that they are perceived as descriptive per se by \emph{all} users.

People's local experiences shape and differentiate their perceptions. The interviews illustrated how tags were interpreted in multiple ways; \texttt{church} was taken to refer to a church building, or to the streetcars servicing the J-Church light-rail line, and was not seen as descriptive by the majority of participants (see Castro in Table~\ref{tab:tagalignment}). Local terms surfaced by our model, such as \texttt{walls} in the Mission, represented the neighborhood's characteristic murals to a long-time local interviewee (M 49), whereas this term was a meaningless for others. While the majority of survey respondents classified \texttt{night} and \texttt{coffee} as unspecific for the Mission, the same interviewee (M 49) for example saw \texttt{night} as descriptive for its bars, restaurants and clubs and thought \texttt{coffee} referred to the copious amounts of coffee shops. Similarly, while one interviewee described North Beach as ``party land'' (M 46), another claimed ``I don't believe there's much of a nightlife there'' (F 26). While these results highlight an opportunity for discovery and recommendation of local content that users may not be aware of, it also means that a careful explanation might be necessary. 

\subsubsection{Model extensions}
\noindent Beyond misunderstanding tags or finding them not specifically descriptive of a certain neighborhood, 
the interviews provided additional clues about the mismatches identified in our survey, as well as individual 
differences. They are organized below in potential opportunities for extensions of the GHM model.

\paragraph*{Sub-region detection} 
According to the majority of our survey participants, 11 tags classified by the GHM as specifically descriptive of North Beach were actually descriptive of another neighborhood. From our interviews, we learned that the terms surfaced by our model for the North Beach neighborhood included references to the bay's waterside and to the tourist attraction Fisherman's Wharf (see for example \texttt{wharf}, \texttt{seal}, \texttt{crab}, \texttt{bay}, \texttt{pier} in Table~\ref{tab:tagalignment}). The locals that responded to our survey (see Table~\ref{tab:tagclass}) and also the interviewees clearly made a distinction between Fisherman's Wharf and North Beach, whereas the set of administrative regions we used for our model did not: according to the data from the city planing department of San Francisco, Fisherman's Wharf is not a neighborhood in itself and is simply a sub-region of North Beach. Clarifications of region borders and detecting emerging sub-neighborhoods (e.g.\ ``as you go further down south it's a totally different neighborhood''), would improve the quality of the neighborhood tags presented to the user. From the modeling perspective, the GHM already allows for considering predefined sub-regions in certain neighborhood, since the geo-tree can be unbalanced and have a different depth for certain sub-trees.

\paragraph*{Permeable adjacency and topography}
Using \textit{permeable adjacency} by considering adjacent neighborhoods appears promising. For example, the mismatching tag \texttt{dolorespark} (see Table~\ref{tab:tagalignment}) refers to the park right on the edge of the Mission and Castro, 
which however was placed outside the Castro neighborhood by the majority of our survey respondents. Interviewees defined neighborhood boundaries differently, and sometimes indicated they didn't know neighborhoods well ``It's a little difficult because it's right next to the Presidio. I kind of, maybe confuse them from each other'' (F 26). Interviewees extended their reasoning about activities or points-of-interest that could spill over into adjacent neighborhoods. For example, the Golden Gate Bridge, officially part of the Presidio neighborhood, but photographed from a wide range of other neighborhoods was mentioned as a distant-but-characteristic feature: ``you can see the Bay Bridge from there'' (M 46). Extensions that consider wider topography and fuzzy boundaries could improve the quality of the results.

\paragraph*{Representative lower levels}
Interviewees at times cited neighborhoods' unique character as representative of wider regional developments, e.g.\ gentrification of neighborhoods, or \ ``California is known for being\ldots very liberal\ldots and it's almost as if a lot it comes from Castro'' (M 35). Hierarchical modeling offers opportunities beyond the identification of locally descriptive content; it could also help find neighborhoods that are representative of a higher level in the geographical hierarchy.

\paragraph*{Temporality}
Time of day, shifting character of a neighborhoods, events, long-term history, and even change itself were referenced by interviewees: ``I think of night, \ldots there's a lot of activity during night time with the bars and the clubs\ldots but before\ldots you wouldn't be caught there at nighttime\ldots 20 years ago it was a different neighborhood.'' (M 49). Since the Flickr tag collection we used spanned multiple years, some aspects of neighborhoods, represented in tags surfaced by our model, that were characteristic at one time but no longer as prominent at present (such as the Halloween celebrations in the Castro neighborhood), were considered out of place (M 32). Yet other interviewees still found such tags characteristic, and freely associated: ``I think of outrageous costumes, and also the costumes that people see when it's not Halloween" (F 26). Combining hierarchical modeling with features detecting diurnal rhythms~\cite{Grinberg:Extract} or events~\cite{Shamma2011} would provide additional insight in such content changes over time. However, we must be aware that this necessitates a dataset that is much more prolific than the one at hand: we need to have a sufficient number of geotagged samples per time period. 

\paragraph*{The non-tagged}
More recent socio-economic developments were mentioned by the interviewees, but not represented in the dataset of tags. For example, one interviewee argued that, while the terms for SoMa did capture the neighborhood (``I think it's really descriptive and pretty accurate"), he would himself add ``something about startups and expensive apartments that really aren't all that great. \$3500 a month for one bedroom. That would be a good descriptor.'' (M 32). Note that even the absence of a feature could be characteristic: ``There's not a lot to do around there." (M 35). Distinguishing however between what is absent in a neighborhood, and what is not represented in a dataset is a challenge~\cite{Rost2013}. Findings proxies for such features, such as proposed by Quercia et al.~\cite{Quercia2014}, requires rigor because it might introduce error and biases. Combining different data sources and model features can however provide additional opportunities.

\section{Discussion and Conclusions}
\label{section:conclusions}
\noindent In this work we proposed a probabilistic model that allows for uncovering terms that are \textit{specifically descriptive} of a region within a given geographical hierarchy. By applying our model to a large-scale dataset of 20 million tags associated with approximately 8 million geotagged Flickr photos taken in San Francisco and Manhattan, we were able to associate each node in the hierarchy to the tags that specifically describe it. Moreover, we used these descriptions to quantify the uniqueness of neighborhoods, and find a mapping between similar but geographically distant neighborhoods. We further conducted interviews and a survey with local residents in order to evaluate the quality of the results given by the GHM and its ability to surface neighborhood-characteristic terms, which span both terms known and unknown to locals. The classification accuracy of GHM, measured with respect to the classification of tags made by locals, provides strong support for the validity of our approach. However, the results of the interviews also highlighted the difference between the performance of a model at classifying community-generated data, and its performance as judged subjectively by an individual. This highlights the importance of taking user subjectivity into account, and the need for explanation and framing. As a consequence, the traditional evaluation of modeling approaches that are fed with user-generated data faces the challenge of both the representation and the subjectivity inherent to vernacular geography. It is important to note that the dataset itself was not formally labeled and rather only contained free-form, often noisy user-generated tags. While unstructured user-generated tags can be cleaned, for instance through canonicalization~\cite{Thomee:2014}, one can utilize other structures inherent in the data to find meaning. Our work takes advantage of the latent geographical patterns exhibited by geotagged data, and shows how we can relate tags to regions within a geographical hierarchy.

\paragraph*{Representation}
We used a specific type of dataset (tags) in this paper to describe geographical regions. We are aware that the tags generated by other online communities that use different social photo sharing services, such as Instagram, may yield different descriptions of regions. For example, if tagging in such a service is more emotive rather than descriptive of the content, our method would surface different local qualities. User-generated data is only representative of the activity of a community within it, and not necessarily of all people in a certain region. Large-scale social media datasets, whether they consist of check-ins, status updates, or photos uploaded to a community-based service, are the result of communicative acts influenced by service design features and evolving community norms~\cite{Rost2013, Shamma2007, 1149949}. In our case, not all local features are captured and shared on Flickr. The content that is present can however be used as relative hints at local trends, and provide comparative insights. Consequently, when we here state locally descriptive, we therefore mean descriptive for the content generated in that locale, not necessarily for all human activity.

\paragraph*{Vernacular geography}
Spatial knowledge that is used to communicate about space and regions has been referred to as vernacular geography~\cite{hollenstein2010exploring} or naive geography~\cite{egenhofer1995naive} that carries with it a certain intrinsic vagueness in its nature~\cite{davies2009user}. Individuals' descriptions of places are inherently subjective, as are interpretations of what is and what is not descriptive for a locale~\cite{Bentley2012draw}. Towards any community, online or offline, the perception of place is considered to be of a shared frame of reference~\cite{montello2003s}. Photo tags generated by the Flickr community are no exception. However, our qualitative exploration is aimed at understanding \textit{what factors come into play} when
describing space, and how might they manifest into a probabilistic model. Fuzzy regional boundaries, temporal events, hidden landmarks all underscore types of regions and terms we surfaced through the interviews, as well as the kinds of tag descriptors discovered quantitatively.  We find a mixed method approach brings a clearer understanding of online communities and naive geography alike.

\paragraph*{Future work}
As illustrated by our generative interviews, the presentation of our model's results ought to take into account subjective interpretations, support the discovery of new conceptual clusters and their explanation. The interviews also pointed to the possibility of using certain neighborhoods as embodying the spirit of its encompassing region, e.g.,\ the Castro described as an influential San Francisco neighborhood, or technology companies moving into SoMa reflecting changes in the city as a whole. The relationship between the leaves of a geographical tree, temporal change and events, sub-regions, and fluid local boundaries that consider wider topographical features represent potential viable extensions to our model, conditioned on the fact that the dataset at hand includes enough data and spans a long period of time. For example, we could extend our model to account for the porosity of frontiers between neighborhoods and the fact that there is no consensus about the exact boundaries of a neighborhood: we define, for each region, a set that is composed of the regions that are adjacent to it. The probability of observing tag in a certain region then includes a new term that accounts for the possibility of sampling tags from adjacent regions. The learning procedure for this extension would be very similar to the GHM's learning procedure.

\end{document}